\documentclass[conference]{IEEEtran}
\IEEEoverridecommandlockouts
\usepackage{float}
\usepackage{amsfonts}
\usepackage{booktabs} 
\usepackage{array}
\usepackage{cite}
\usepackage{amsmath,amssymb,amsfonts}
\usepackage{algorithmic}
\usepackage{graphicx}
\usepackage{tabularx}
\usepackage{subcaption}
\usepackage{physics}
\usepackage{url}
\usepackage{physics}
\usepackage{placeins}
\usepackage{graphicx,tikz,amsmath}
\usetikzlibrary{arrows.meta,positioning,calc,fit,backgrounds}

\usepackage{textcomp}
\usepackage{xcolor}
\usepackage{newunicodechar}
\usepackage{siunitx}
\usepackage{tikz}
\usetikzlibrary{positioning, fit, backgrounds, arrows.meta, calc, shapes.geometric}

\newunicodechar{−}{-}
\def\BibTeX{{\rm B\kern-.05em{\sc i\kern-.025em b}\kern-.08em
    T\kern-.1667em\lower.7ex\hbox{E}\kern-.125emX}}
\begin{document}

\title{Tilt-Rotor Quadcopter with Learned Residual Control and Mode-Dependent State Estimation

\author{\IEEEauthorblockN{Ninad D. Joshi}
\IEEEauthorblockA{\textit{Mechanical and Aerospace Engineering} \\
\textit{Indian Institute of Technology Hyderabad}\\
Hyderabad, India \\
ninadjoshi2002@gmail.com}
\and

\IEEEauthorblockN{Prabhat Kumar}
\IEEEauthorblockA{\textit{Mechanical and Aerospace Engineering} \\
\textit{Indian Institute of Technology Hyderabad}\\
Hyderabad, India \\
pkumar.mae@iith.ac.in}\and


}
}
\maketitle

\begin{abstract}
Tilt-rotor quadcopters redirect thrust for translation without banking, but complicate modeling, state estimation, and disturbance rejection. This paper presents a simulation-based flight stack for a \SI{2.00}{\kilo\gram} tilt-rotor quadcopter. The system uses a six-degree-of-freedom nonlinear model with first-order motor dynamics and two-axis rotor tilt, parameterized using a 3D CAD model. A cascaded PD controller commands forward and lateral tilt angles. A 14-state EKF fuses sensor data across hover, early-transition, and transition modes using a hysteretic (Schmitt-trigger) classifier to prevent covariance discontinuities. Because the proportional position loop leaves a steady-state offset under constant wind, a small bounded neural residual is added on top of the unmodified PD controller. It is trained by behavioral cloning from a teacher with privileged access to the true wind, while the deployed network never observes wind. In simulation, the EKF achieves a post-settling position root mean square error (RMSE) of \SI{0.226}{\metre} (horizontal) and \SI{0.03}{\metre} (vertical) and an attitude RMSE below \SI{0.8}{\degree}, and it estimates a wind velocity of $(\num{2.0}, \num{-1.2})\,\unit{\metre\per\second}$ to within \SI{0.15}{\metre\per\second} per axis. The residual reduces steady-state station-keeping error under wind by approximately \SI{49}{\percent}, from \SI{0.117}{\metre} to \SI{0.060}{\metre} at \SI{1.5}{\metre\per\second} and from \SI{0.319}{\metre} to \SI{0.162}{\metre} at an out-of-distribution \SI{3.9}{\metre\per\second}. The framework provides a basis for hardware-in-the-loop and flight-test studies.

\end{abstract}


\section{Introduction}
Unmanned aerial vehicles (UAVs) have gained significant attention due to their wide range of applications in videography, agriculture, surveying, construction, education, and research. Conventional multirotor UAVs provide excellent hovering capability and maneuverability, but their rotors are fixed relative to the airframe. The vehicle is therefore underactuated, and translational motion must be produced by tilting the whole vehicle through changes in motor speed.

A tilt-rotor quadcopter adds actuation by allowing the rotor assemblies to change their orientation \cite{SU2022107561}. The thrust vector can then be redirected to generate translational force directly, rather than relying solely on changes in rotor thrust magnitude, at the cost of stronger coupling between the translational and rotational dynamics. The configuration considered in this work, the vehicle employs twelve actuator inputs: four motor inputs for rotor thrust, four actuators for tilting along the $\alpha$ direction, and four actuators for tilting along the $\beta$ direction. The nominal thrust of the four rotors is assumed to remain constant, and the required $\alpha$ and $\beta$ angles are determined from the desired trajectory. The vehicle motion is simulated using a position-based trajectory-tracking approach rather than a prescribed flight duration. The simulation continues until the vehicle reaches the final waypoint, so the completion time emerges from the vehicle dynamics and control response. A time budget terminates runs that fail to converge to a commanded waypoint.
\begin{figure}[H]
    \centering
    \includegraphics[width=0.8\columnwidth]{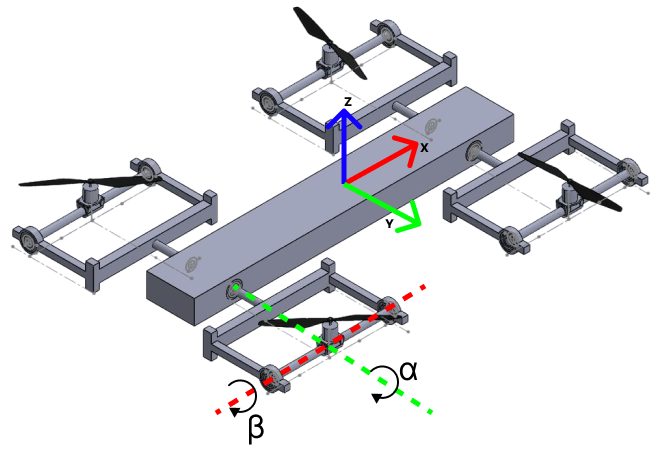}
    \caption{Orientation along with tilt-rotation representation of $\alpha$ and $\beta$ which produce forward and lateral force components respectively.}
    \label{fig:drone_design}
\end{figure}

Existing studies often employ simplified mass and inertia parameters along with simple control strategies, whereas the influence of wind and complex trajectories on the resulting dynamic response has been less extensively investigated \cite{6564796, Elfeky2015,  AnhTa2012}. Previous work has implemented standard PID controllers \cite{DONADEL20143839, 6842367, Liang}. 

This work instead develops a six-degree-of-freedom nonlinear dynamic model of a tilt-rotor UAV using the physical and inertial parameters obtained from its 3D CAD model, and evaluates its trajectory and attitude response. For control, it adds a neural network (NN) that acts as a small additive correction on a PD controller.
Our goal is to study whether a bounded learned correction can recover the
wind-trim offset of the proportional loop; we do not claim superiority over
integral action or disturbance observers, which are not evaluated here.

\section{Mathematical Model of the Tilt-Rotor Quadcopter}
\subsection{State and Notation}

The vehicle state is
\begin{equation}
\small
\vec{x} = \Big[\ \underbrace{x,y,z}_{\text{position}},\
\underbrace{u,v,w}_{\text{body velocity}},\
\underbrace{\phi,\theta,\psi}_{\text{attitude}},\
\underbrace{p,q,r}_{\text{body rate}},\
\underbrace{T_1,T_2,T_3,T_4}_{\text{rotor thrust}}\ \Big]^{\!\top}
\in \mathbb{R}^{16},
\end{equation}
where $(x,y,z)$ is position in the world (inertial) frame, $(u,v,w)$ is
linear velocity expressed in the \emph{body} frame, $(\phi,\theta,\psi)$
are the roll--pitch--yaw Euler angles, $(p,q,r)$ are the body angular
rates, and $T_i$ is the actual (lagged) thrust of rotor~$i$. The vehicle
has mass $m$, gravity is $g$, and the inertia tensor is
$\mathrm{diag}(I_{xx},I_{yy},I_{zz})$. Attitude uses the ZYX Euler
sequence, $\vec{R}(\phi,\theta,\psi)=\vec{R}_z(\psi)\vec{R}_y(\theta)\vec{R}_x(\phi)$,
mapping body to world coordinates. Aerodynamic drag is linear in the air-relative velocity, {$F_{\text{drag}} = -\mathrm{diag}(k_u,k_v,k_w)(v_b - R^{\top}v_{\text{wind}})$, and is added to $F_{\text{body}}$}. The rigid-body translational and
rotational dynamics, the Euler-angle kinematics, and the first-order
motor/ESC lag, $\dot T_i=(T_{c,i}-T_i)/\tau_m$, follow the standard
quadcopter formulation \cite{Beard2008,Mahony2012,Stevens2015}. 

\subsection{Tilt-Rotor Actuation and Allocation}

Each rotor is mounted on a two-axis gimbal providing a forward tilt
$\alpha_i$ (rotation about body $Y$) and a lateral tilt $\beta_i$
(rotation about body $X$). With the nominal thrust axis along body $Z$,
the unit thrust direction is
$\vec{d}_i=\vec{R}_x(-\beta_i)\vec{R}_y(\alpha_i)[0\ 0\ 1]^{\top}$, so
the thrust and reaction torque of rotor $i$ (magnitude $T_i$) are
\begin{equation}
\vec{F}_i = T_i\,\vec{d}_i,\quad
\vec{d}_i=\begin{bmatrix}
\sin\alpha_i\cos\beta_i\\ \sin\beta_i\\ \cos\alpha_i\cos\beta_i
\end{bmatrix},\quad
\vec{\tau}_{r,i}=s_i\,k\,T_i\,\vec{d}_i,
\label{eq:tilt}
\end{equation}
where $s_i\in\{+1,-1\}$ is the spin direction and $k$ is the
torque-to-thrust ratio. Positive $\alpha$ therefore produces a  forward force component (+X) and positive $\beta$ produces a  lateral component (+Y). With rotor $i$ at body-fixed position $\vec{r}_i$ relative to the center of mass, the net force and moment on the airframe are
\begin{equation}
\vec{F}_{\mathrm{body}}=\sum_{i=1}^{4}\vec{F}_i,\qquad
\vec{\tau}_{\mathrm{body}}=\sum_{i=1}^{4}\big(\vec{r}_i\times\vec{F}_i+\vec{\tau}_{r,i}\big).
\label{eq:allocation}
\end{equation}
For a common tilt ($\alpha_i\equiv\alpha$, $\beta_i\equiv\beta$), equal
thrust, symmetric placement ($\sum_i\vec{r}_i=\vec{0}$), and balanced
spin directions ($\sum_i s_i=0$), the net moment vanishes, so the vehicle
translates through tilt alone without induced roll, pitch, or yaw torque.

\subsection{Sensor Models}

Six sensors are modeled, each with additive Gaussian noise and a constant
bias: a \SI{100}{\hertz} gyroscope and accelerometer (specific force, thrust
approximated as acting along body $Z$), a \SI{5}{\hertz} GPS receiver (world
position), a \SI{20}{\hertz} barometer (altitude), a \SI{20}{\hertz} three-axis
magnetometer (measuring a fixed-inclination world field vector rotated
into the body frame), and a \SI{10}{\hertz} pitot tube measuring true airspeed
$\lVert\mathbf v_b - R^\top\mathbf v_{\mathrm{wind}}\rVert$.

\section{Control and State Estimation}
\label{sec:control}
\label{sec:ekf}

\subsection{Cascaded control}
\begin{figure}[t]
\centering
\resizebox{0.35\textwidth}{!}{%
\begin{tikzpicture}[
    >=stealth,
    node distance=0.45cm and 0.8cm,
    font=\sffamily\small,
    block/.style={
        draw=black, fill=white, rounded corners=4pt,
        minimum width=2.4cm, minimum height=0.75cm, align=center, inner sep=4pt
    }
]%
    \node [block] (pos) {Position\\Controller};%
    \node [block, below=of pos] (vel) {Velocity\\Controller};%
    \node [block, below=of vel] (att) {Attitude\\Controller};%
    \node [block, below=of att] (rate) {Rate\\Controller};%
    \node [font=\sffamily\small, below=0.15cm of rate] (pid_label) {Cascaded PID};%

    \begin{scope}[on background layer]%
        \node [draw=black, solid, thick, fill=white, rounded corners=2pt,
               fit={(pos) (pid_label)}, inner sep=6pt] (pid_box) {};%
    \end{scope}%

    \node [block, right=0.8cm of pos] (features) {15 Input\\Features (z-\\scored)};%
    \node [block, below=0.4cm of features] (mlp) {MLP\\(15-64-64-6)};%
    \node [block, below=0.4cm of mlp] (tanh) {Tanh activation\\(Bounded\\outputs)};%
    \begin{scope}[on background layer]%
        \node [draw=black, dashed, thick, fill=white, rounded corners=2pt,
               fit={(features) (tanh)}, inner sep=6pt] (nn_box) {};%
    \end{scope}%

    \node [block, right=0.8cm of mlp.east] (mixer) {Inverse\\Mixer};%
    \node [block, below=0.4cm of mixer] (lag1) {Motor\\Lag};%
    \node [block, below=0.4cm of lag1] (lag2) {Motor\\Lag};%
    \begin{scope}[on background layer]%
        \node [draw=black, dashed, thick, fill=white, rounded corners=2pt,
               fit={(mixer) (lag2)}, inner sep=6pt] (act_box) {};%
    \end{scope}%

    \node [block, right=0.8cm of lag2] (rot_dyn) {Rotational\\Dynamics};%
    \node [block, above=0.3cm of rot_dyn] (trans_dyn) {Translation\\Dynamics};%
    \node [font=\sffamily\small, below=0.15cm of rot_dyn] (quad_label) {Quad Dynamics};%

    \begin{scope}[on background layer]%
        \node [draw=black, solid, thick, fill=white, rounded corners=2pt,
               fit={(trans_dyn) (quad_label)}, inner sep=6pt] (quad_box) {};%
    \end{scope}%

    \node [block, above=0.6cm of quad_box] (detector) {Flight mode\\detector};%
    \node [block, above=0.6cm of detector] (ekf) {14 state\\estimate(EKF)};%
    \node [block, above=0.6cm of ekf] (predict) {Predict(RK4)};%
    \node [block, above=0.6cm of predict] (imu) {IMU\\(100Hz, every cycle)};%

    \node [block, right=1.0cm of imu] (aiding) {Aiding Sensors (GPS,\\Baro, Mag, Airspeed)};%
    \node [block] at (aiding |- ekf) (mode_update) {Mode-gated\\update};%

    \begin{scope}[on background layer]%
        \node [draw=black, dashed, thick, fill=white, rounded corners=2pt,
               fit={(imu) (aiding)}, inner sep=6pt] (sensor_box) {};%
    \end{scope}%
    \begin{scope}[on background layer]%
        \node [draw=black, dashed, thick, fill=white, rounded corners=2pt,
               fit={(predict) (ekf) (mode_update)}, inner sep=6pt] (ekf_box) {};%
    \end{scope}%

    \draw [->, thick] (pos) -- (vel);%
    \draw [->, thick] (vel) -- (att);%
    \draw [->, thick] (att) -- (rate);%
    \draw [->, thick] (pos.east) -- (features.west);%
    \draw [->, thick] (features) -- (mlp);%
    \draw [->, thick] (mlp) -- (tanh);%
    \draw [->, thick] (tanh.south) -- ++(0,-0.4) -| ($(act_box.west)+(-0.3,0)$) |- (mixer.west);%
    \draw [->, thick] (mixer) -- (lag1);%
    \draw [->, thick] (lag1) -- (lag2);%
    \draw [->, thick] (act_box.east |- lag2) -- (quad_box.west |- lag2);%
    \draw [->, thick] (quad_box.north) -- (detector.south);%
    \draw [->, thick] (ekf.south) -- (detector.north);%
    \draw [->, thick] (imu) -- (predict);%
    \draw [->, thick] (predict) -- (ekf);%
    \draw [->, thick] (detector.east) -| (mode_update.south);%
    
    \draw [->, thick] (aiding) -- (mode_update);%
    \draw [->, thick] (mode_update) -- (ekf);%

    \draw [->, thick] (quad_box.east) -- ++(0.3,0) |- ($(pid_box.south)+(0,-0.4)$) -| ($(pid_box.west)+(-0.5,0)$) |- (pid_box.west |- att);%
\end{tikzpicture}%
}
\caption{Control Architecture}
\label{fig:cascade}
\end{figure}
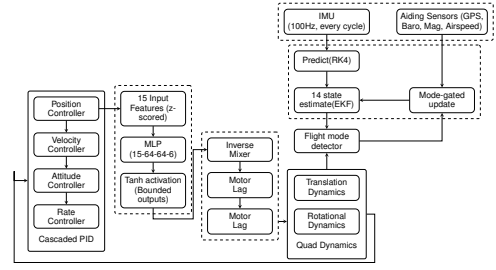
The controller follows the standard multirotor cascade of fast inner rate and
attitude loops under slower velocity and position loops
(Fig.~\ref{fig:cascade}) \cite{Mahony2012,Bouabdallah2007}, so each loop can be
tuned assuming the inner loop has settled. The horizontal position error is
rotated into the body frame by the current yaw,
$\mathbf e_b=R_\psi^\top(\mathbf p_d-\mathbf p)_{xy}$, and a proportional law
gives the desired body-frame velocity $u_d=K_{p,\mathrm{pos}}\,e_{b,x}$ (and
analogously $v_d$), saturated to $\pm v_{\max}$. Altitude uses a PD law with
gravity feed-forward,
$F=mg+K_{p,z}(z_d-z)-K_{d,z}\dot z$, divided by
$\max(\cos\phi\cos\theta,0.5)$. Unlike a fixed-rotor quadcopter, the
horizontal velocity error is mapped to tilt angles instead of a roll/pitch
command; inverting~\eqref{eq:tilt} at small angle gives
\begin{equation}
\qquad 
\alpha = \arctan\!\Big(\frac{\eta\,K_{p,\alpha}(u_d-u)\cos\beta}{F}\Big)
\label{eq:servo_alloc}
\end{equation}
\begin{equation}
\beta = \arctan\!\Big(\frac{K_{p,\beta}(v_d-v)}{F}\Big)
\end{equation}
{where} $\eta=\max(0,\cos\psi_e)$ suppresses forward thrust when the heading
error $\psi_e$ is large, and both angles are clipped to $\pm\alpha_{\max}$.
Attitude errors drive a proportional law to desired body rates, and a PD law
on rate error (derivative taken from a low-pass-filtered rate) produces the
torque command $\boldsymbol\tau$. Given $U=[F\ \tau_x\ \tau_y\ \tau_z]^\top$
and the current servo angles, an effectiveness matrix $A\in\mathbb R^{4\times4}$
is built from the per-motor force/torque map at unit thrust, and the rotor
thrusts follow from $A\mathbf T=U$ with a least-squares fallback near
ill-conditioning, a standard control-allocation formulation
\cite{Johansen2013} used with tilting rotors in
\cite{Ryll2015,6564796,DONADEL20143839}.

\subsection{Estimator}
The EKF follows the usual INS/GNSS structure \cite{Groves2013}, with state
$\hat{\mathbf x}=[\mathbf p_w^\top,\mathbf v_w^\top,\phi,\theta,\psi,\mathbf b_g^\top,w_n,w_e]^\top\in\mathbb R^{14}$
(world-frame position and velocity, attitude, gyro bias, horizontal wind).
Bias-corrected gyro rates propagate the attitude through the standard Euler
kinematics \cite{Beard2008}. Velocity follows a constant-velocity model
$\dot{\mathbf v}_w=\mathbf 0+\text{noise}$ instead of integrating the
accelerometer, because the specific force depends on the servo angles
$\alpha,\beta$, which are not filter states. The model is integrated with a
fourth-order Runge--Kutta step, and the covariance is propagated with
$\Phi\approx I+A\Delta t+\tfrac12(A\Delta t)^2$ using a central-difference
Jacobian $A$. GPS and barometer updates are linear. As in gravity- and
field-aided attitude estimation \cite{Crassidis2007}, the accelerometer
provides only a roll/pitch measurement $h(\hat{\mathbf x})=R^\top[0\ 0\ g]^\top$,
gated by a quasi-static test on $\lvert\lVert\mathbf a\rVert-g\rvert$, and the
tilt-compensated magnetometer provides a scalar heading measurement against
$\psi$ with a wrapped innovation. Following GPS/airspeed wind estimation for
small UAVs \cite{Langelaan2011,Waslander2009}, the pitot measurement is fused
jointly against velocity and wind,
$h(\hat{\mathbf x})=\lVert\mathbf v_w-[w_n\ w_e\ 0]^\top\rVert$: GPS
constrains only ground velocity and airspeed only the speed relative to the
air, so together, under changing heading, they make wind observable.

\subsection{Mode-dependent sensor set.}
At every filter cycle a three-state machine selects the fused sensors from
the estimated airspeed $\hat V=\lVert\hat{\mathbf v}_w-\hat{\mathbf w}_h\rVert$
(not the raw pitot sample, so one outlier cannot flip the mode) and the
commanded tilt $\lvert\alpha\rvert$. Each boundary has separate rising and
falling thresholds (Schmitt trigger), so a signal near a boundary does not
toggle the mode and abruptly change which measurements the filter trusts.
Table~\ref{tab:modes} lists the sensor sets and switching conditions.

\begin{table}[t]
\centering
\caption{Mode-dependent sensor sets and Schmitt-trigger switching conditions}
\label{tab:modes}
    \centering
    \includegraphics[width=\columnwidth]{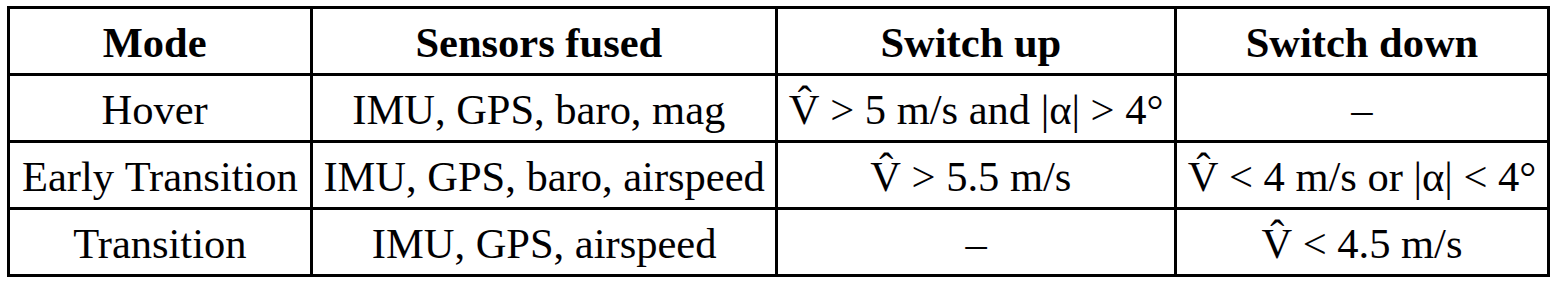}
\end{table}

\section{Neural Residual Controller}
\label{sec:nn}

The position loop of Section~\ref{sec:control} is purely proportional, so a
constant wind produces a nonzero steady-state position error by
construction. Rather than altering the validated PD cascade, we leave every
equation of Section~\ref{sec:control} unchanged and add a bounded learned
correction. Let $(F,\alpha,\beta,\boldsymbol\tau)$ be the PD commands. A
function $r(\cdot)$ returns $(dF,d\alpha,d\beta,d\boldsymbol\tau)$, applied as
\begin{equation}
\begin{aligned}
F &\leftarrow \max(F+dF,F_{\min}), & \alpha &\leftarrow \mathrm{clip}(\alpha+d\alpha),\\
\beta &\leftarrow \mathrm{clip}(\beta+d\beta), & \boldsymbol\tau &\leftarrow \boldsymbol\tau+d\boldsymbol\tau,
\end{aligned}
\label{eq:residual_hook}
\end{equation}
with the same clip bounds the PD path already enforces, so the residual
cannot command the actuators beyond their limits (this bounds the command,
not closed-loop stability). $r(\cdot)$ is an MLP with a 15-dimensional input,
two 64-unit ReLU hidden layers, and a 6-dimensional $\tanh$ output scaled by
fixed limits (thrust: \SI{15}{\percent} of hover thrust; each tilt angle:
\SI{5}{\degree}; each torque channel: \SI{0.05}{\newton\metre}). The output
layer is zero-initialized, so the combined controller equals the PD baseline
at the start of training.

The 15 inputs use only quantities the PD cascade already computes:
body-frame position and velocity error, roll, pitch, the sine and cosine of
the heading error, body rates, and the PD loop's $(\alpha,\beta,F)$. The
network never receives the wind. Under proportional control a constant wind
yields a repeatable error signature, which the network can map to a trim
correction, much as an integral term compensates a disturbance it never
measures.

{Training labels come from a privileged teacher with access to the true wind, used only during data generation. From the drag model
$F_{\text{drag}} = -\mathrm{diag}(k_u,k_v,k_w)(v_b - R^\top v_{\text{wind}})$
and $F_x\!\approx\!F\sin\alpha$ (Eq.~\eqref{eq:tilt}, $\cos\beta\!\approx\!1$), the trim correction is}
\begin{equation}
d\alpha_{\mathrm{ff}} = \mathrm{atan2}\big(-k_u\, w^b_{\mathrm{wind},x},\,F\big),
\label{eq:teacher}
\end{equation}
and symmetrically for $d\beta_{\mathrm{ff}}$ with $k_v$ and
$w^b_{\mathrm{wind},y}$. Thrust and torque targets are zero.
Equation~\eqref{eq:teacher} could be applied directly as a feedforward; we use
it as a label so that the pipeline extends to teachers without a closed form.
Training is single-round behavioral cloning (not DAgger) on \num{25621} samples
from \num{40} PD-only rollouts with random wind
(\SIrange{0}{3}{\metre\per\second}, uniform direction)
and random missions, using Adam on mean-squared error with feature
standardization from the training split. Analytic gradients were verified
against central finite differences (maximum relative error below \num{1e-3}).

\section{Results}
\label{sec:results}

\subsection{Setup}

{The vehicle ($m=\SI{2.00}{\kilo\gram}$, arm length
\SI{0.225}{\metre},
$\mathbf{I}=\operatorname{diag}(\num{0.01758},\,\num{0.01968},\,\num{0.02471})\,\si{\kilo\gram\metre\squared}$,
$\alpha_{\max}=\beta_{\max}=\SI{20}{\degree}$) flies a six-waypoint
mission (climb to \SI{10}{\metre}, traverse a \SI{20}{\metre} square,
and land), with a \SI{0.1}{\metre} capture radius and a
\SI{100}{\hertz} estimator.} The filter starts with a
\SI{1.5}{\metre} position error and a \SI{7.02}{\degree} attitude
error so that convergence is visible. Because the capture event stops
a leg at \SI{0.1}{\metre} and masks the true steady-state offset,
controller comparisons use a station-keeping test: the vehicle flies
to a setpoint and holds for \SI{30}{\second}, and the mean horizontal
error over the second half is reported.
\begin{figure}[t]
\centering
\begin{subfigure}[t]{0.16\textwidth}
  \includegraphics[width=\linewidth]{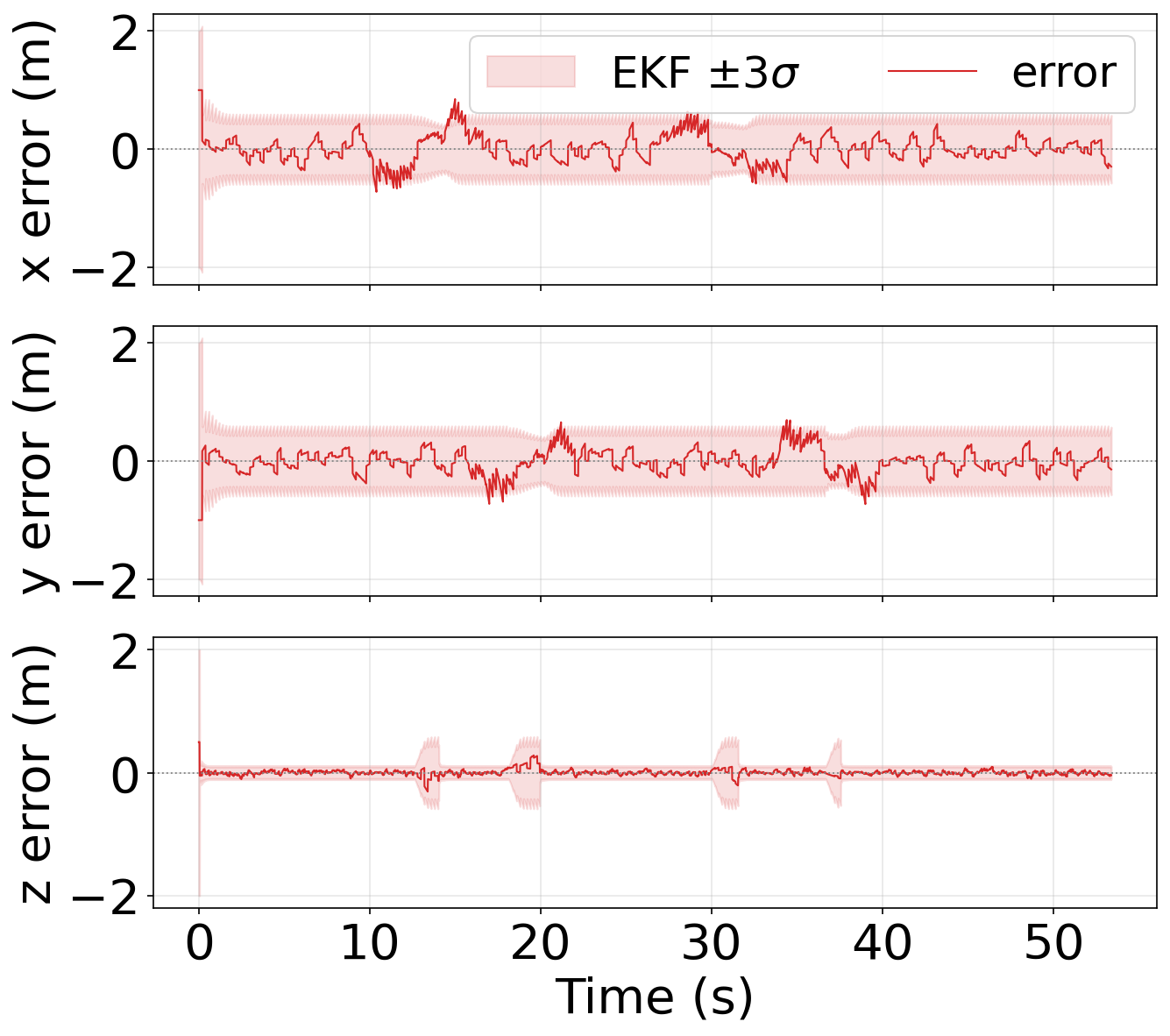}
  \caption{}
\end{subfigure}\hfill
\begin{subfigure}[t]{0.16\textwidth}
  \includegraphics[width=\linewidth]{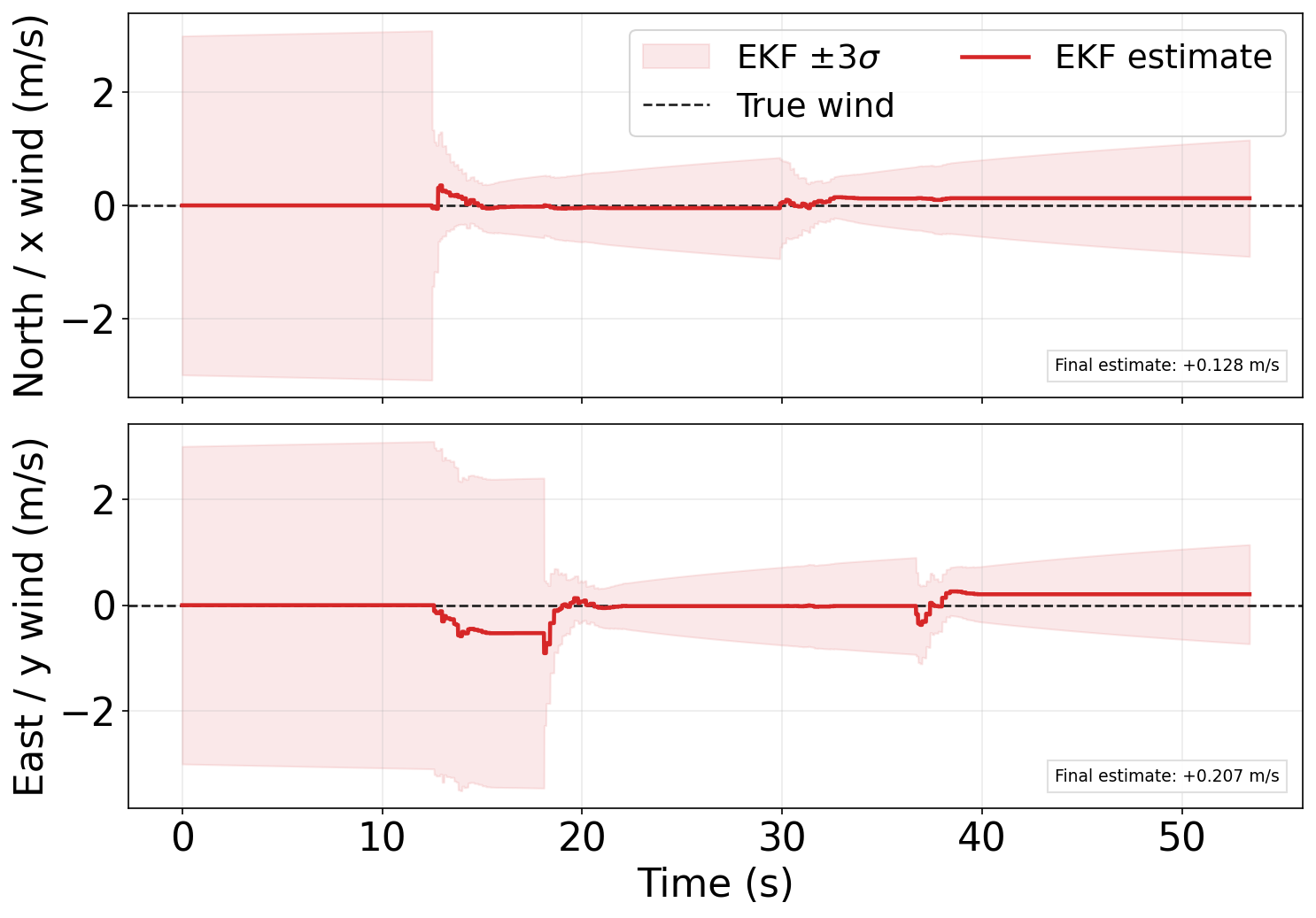}
    \caption{}

\end{subfigure}\hfill 
\begin{subfigure}[t]{0.16\textwidth}
  \includegraphics[width=\linewidth]{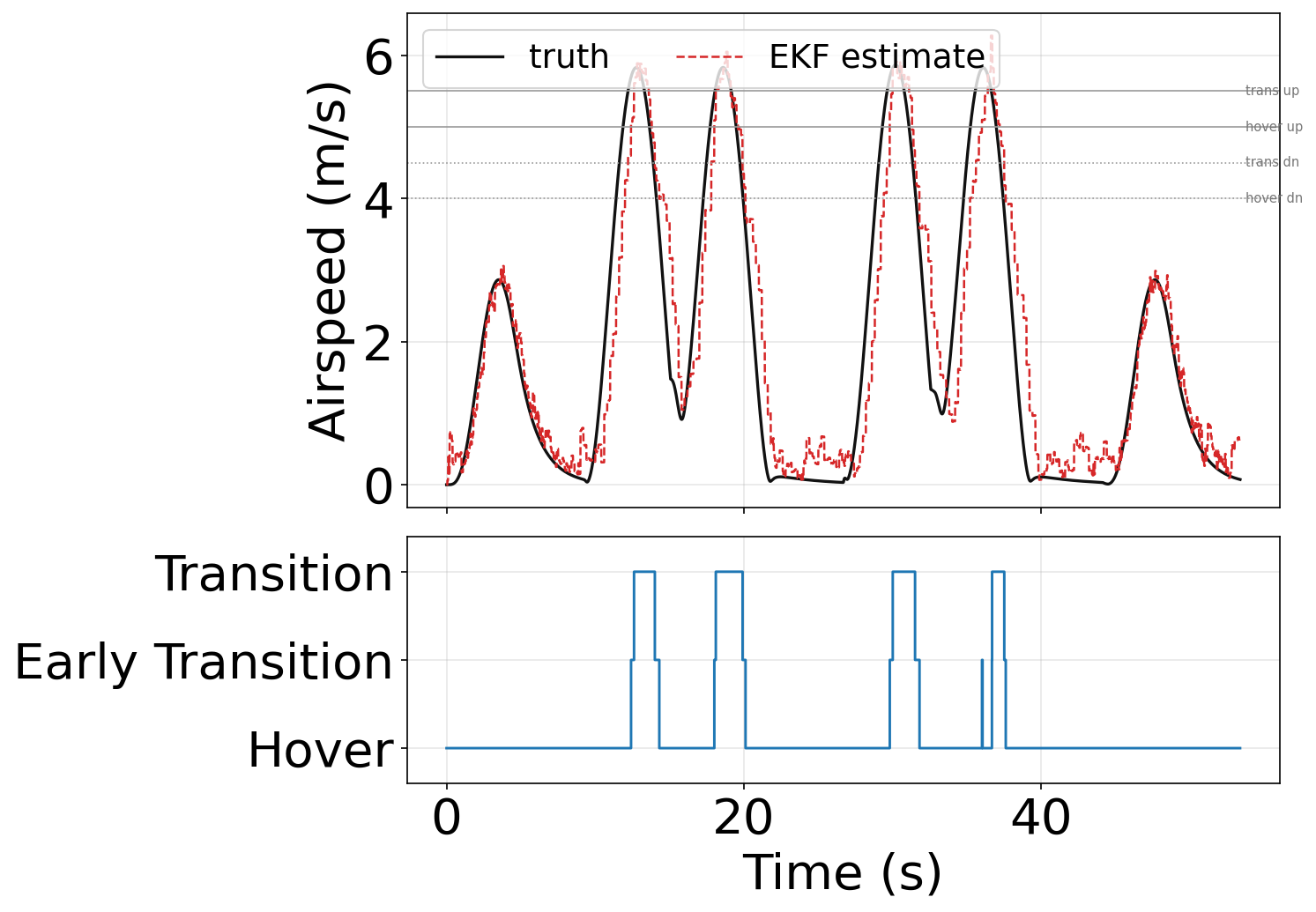}
    \caption{}

\end{subfigure}
\caption{(a) position error against the filter's $\pm3\sigma$
bound, (b) wind estimated in Transition mode, and (c) estimated airspeed with
the Schmitt-trigger mode sequence.}
\label{fig:est}
\end{figure}
\subsection{State Estimation and Mode Switching}

After a \SI{5}{\second} settling window, the position RMSE is
\SI{0.226}{\metre}, \SI{0.20}{\metre}, and \SI{0.049}{\metre}
($x$, $y$, $z$), the velocity RMSE is
\SI{0.568}{\metre\per\second}, \SI{0.584}{\metre\per\second}, and
\SI{0.143}{\metre\per\second}, and the attitude RMSE is
\SI{0.439}{\degree}, \SI{1.5}{\degree}, and \SI{1.931}{\degree}
(roll, pitch, yaw). The position error falls below \SI{1}{\metre} at
\SI{0.20}{\second} and the attitude error below \SI{5}{\degree} at
\SI{0.05}{\second}. The gyro bias, which has no dedicated sensor, is
recovered to within \SI{0.6}{\milli\radian\per\second} per axis.
The mean NIS is \num{3.24} for GPS (expected \num{3}) and \num{2.00}
for the accelerometer (expected \num{3}, indicating a conservative
noise model); the accelerometer gate accepted \SI{99.6}{\percent}
of samples (\num{7098}/\num{7124}).
Fig.~\ref{fig:est}(a) shows that the position error remains inside the
$\pm3\sigma$ bound.

In a second mission (\SI{80}{\metre} legs, cruise
\SI{14}{\metre\per\second}, wind
$(\num{2.0},\num{-1.2})\,\si{\metre\per\second}$),
the vehicle spent \SI{69.7}{\percent} of the time in Hover,
\SI{8.8}{\percent} in Early Transition, and \SI{21.5}{\percent} in
Transition (Fig.~\ref{fig:est}(c)). The wind was estimated as
$(\num{1.859},\num{-1.208})\,\si{\metre\per\second}$
against the true
$(\num{2.000},\num{-1.200})\,\si{\metre\per\second}$
(Fig.~\ref{fig:est}(b)), which is observable because Transition mode
fuses GPS and airspeed jointly. Three of the six legs missed the
capture radius (miss distances \SIrange{0.188}{0.201}{\metre}),
consistent with the steady-state offset that
Section~\ref{sec:nn} targets.

\subsection{Residual Controller Performance}

\begin{figure}[t]
\centering
\begin{subfigure}[t]{0.16\textwidth}
  \includegraphics[width=\linewidth]{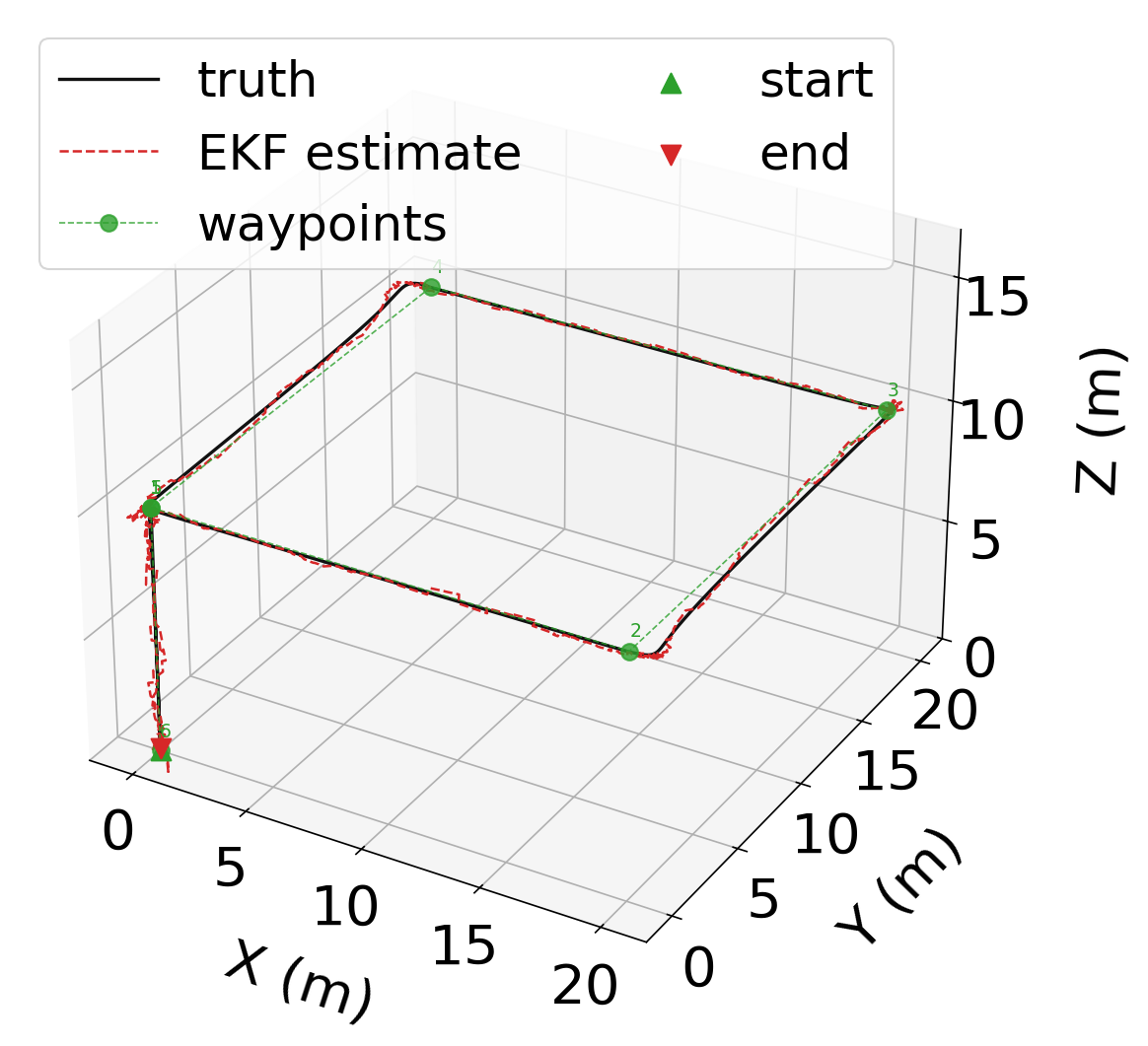}
    \caption{}

\end{subfigure}\hfill
\begin{subfigure}[t]{0.16\textwidth}
  \includegraphics[width=\linewidth]{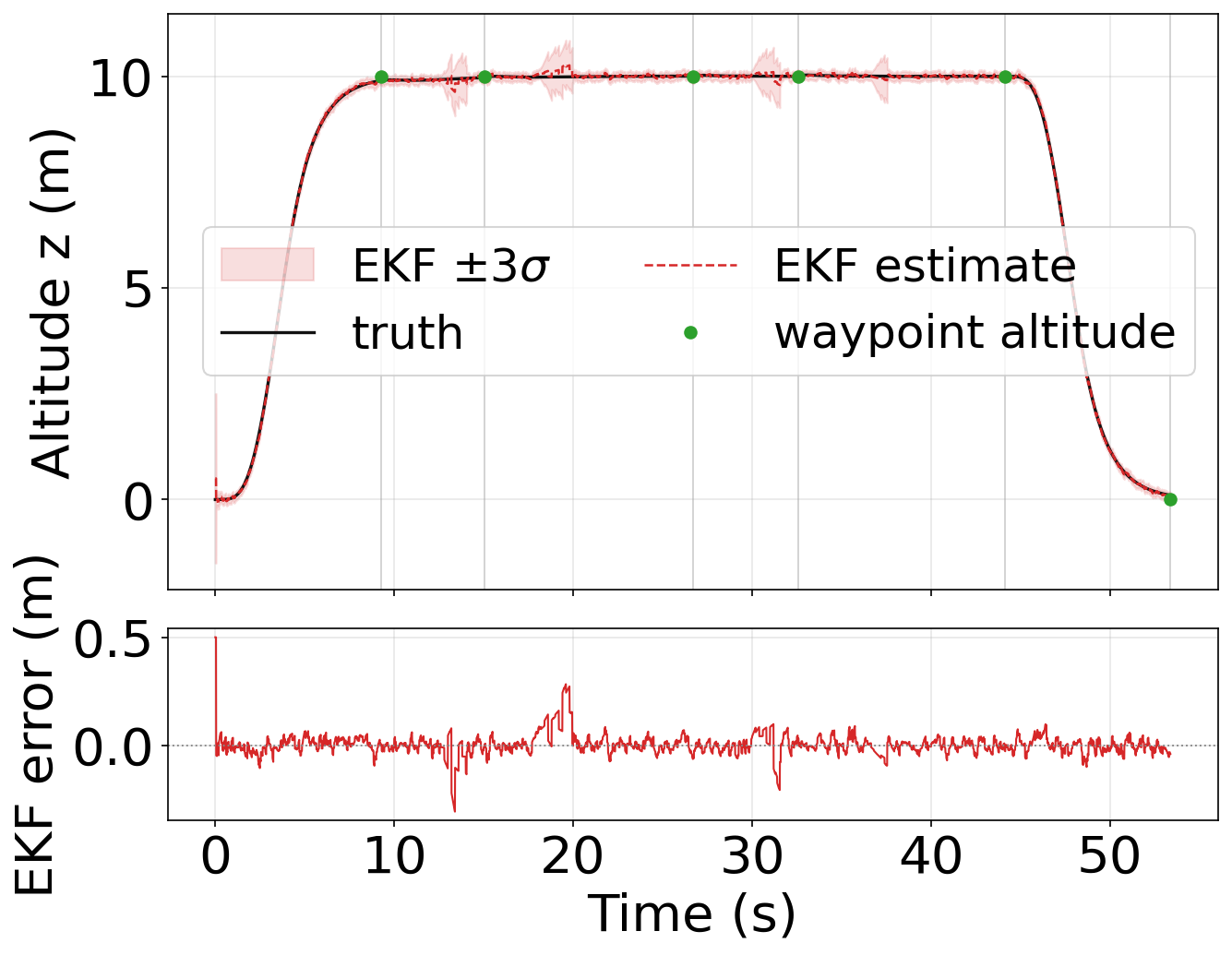}
    \caption{}

\end{subfigure}\hfill
\begin{subfigure}[t]{0.16\textwidth}
  \includegraphics[width=\linewidth]{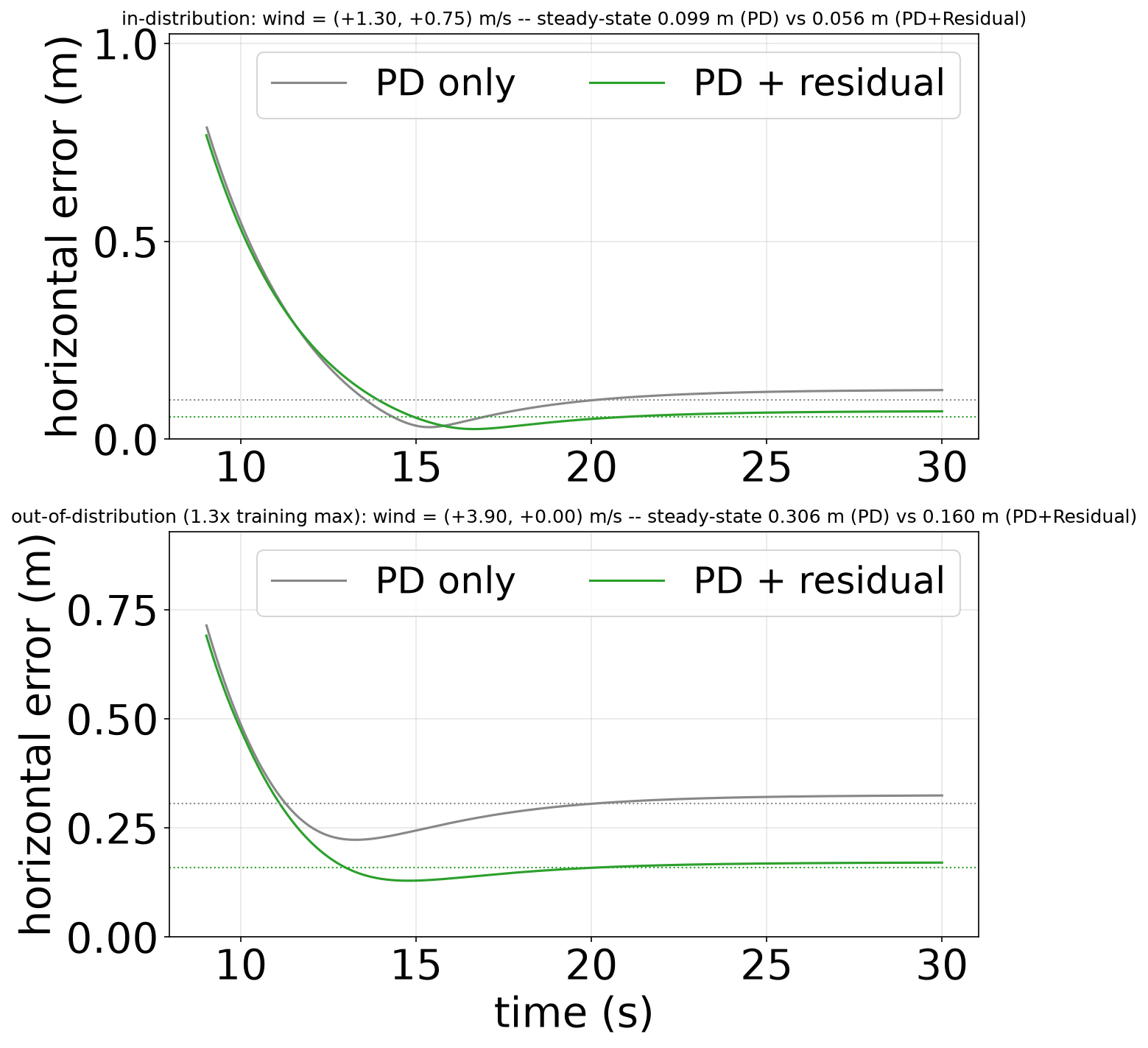}
    \caption{}

\end{subfigure}\hfill

\caption{(a) Flight trajectory against the commanded waypoints, (b) altitude and (c) horizontal error while holding station, PD only vs.\ PD+residual, at an in-distribution and an out-of-distribution wind speed.}
\label{fig:traj1}
\end{figure}

Table~\ref{tab:holdstation} reports the station-keeping error under wind.
The residual reduces the error by \SI{43.9}{\percent} at
\SI{1.5}{\metre\per\second} (within the training range) and by
\SI{47.9}{\percent} at \SI{3.9}{\metre\per\second}, which is
\SI{30}{\percent} beyond the \SI{3.0}{\metre\per\second} training
ceiling (Fig.~\ref{fig:traj1}(c)). The near-identical reduction is
consistent with, though not proof of, the network having learned an
approximation of the wind-trim relationship rather than a lookup over
training conditions. The thrust and torque output channels carry near-zero
learned weight, matching their all-zero teacher targets. Both servo angles
remain within their \SI{\pm 20}{\degree} limits
(Fig.~\ref{fig:traj1}(b)-(c)).

\begin{figure}[t]
\centering

\begin{subfigure}[t]{0.16\textwidth}
  \includegraphics[width=\linewidth]{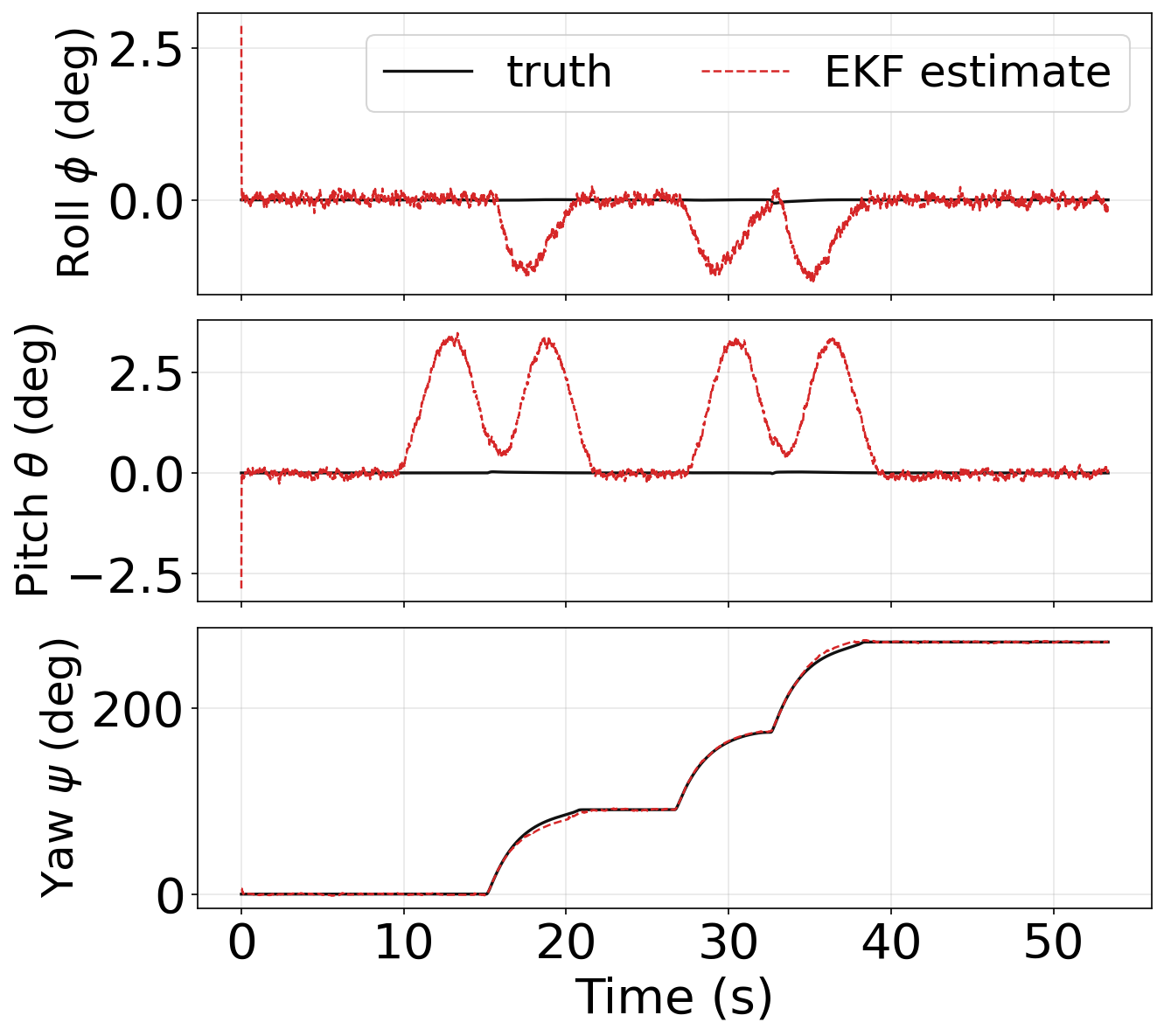}
    \caption{}

\end{subfigure}\hfill
\begin{subfigure}[t]{0.16\textwidth}
  \includegraphics[width=\linewidth]{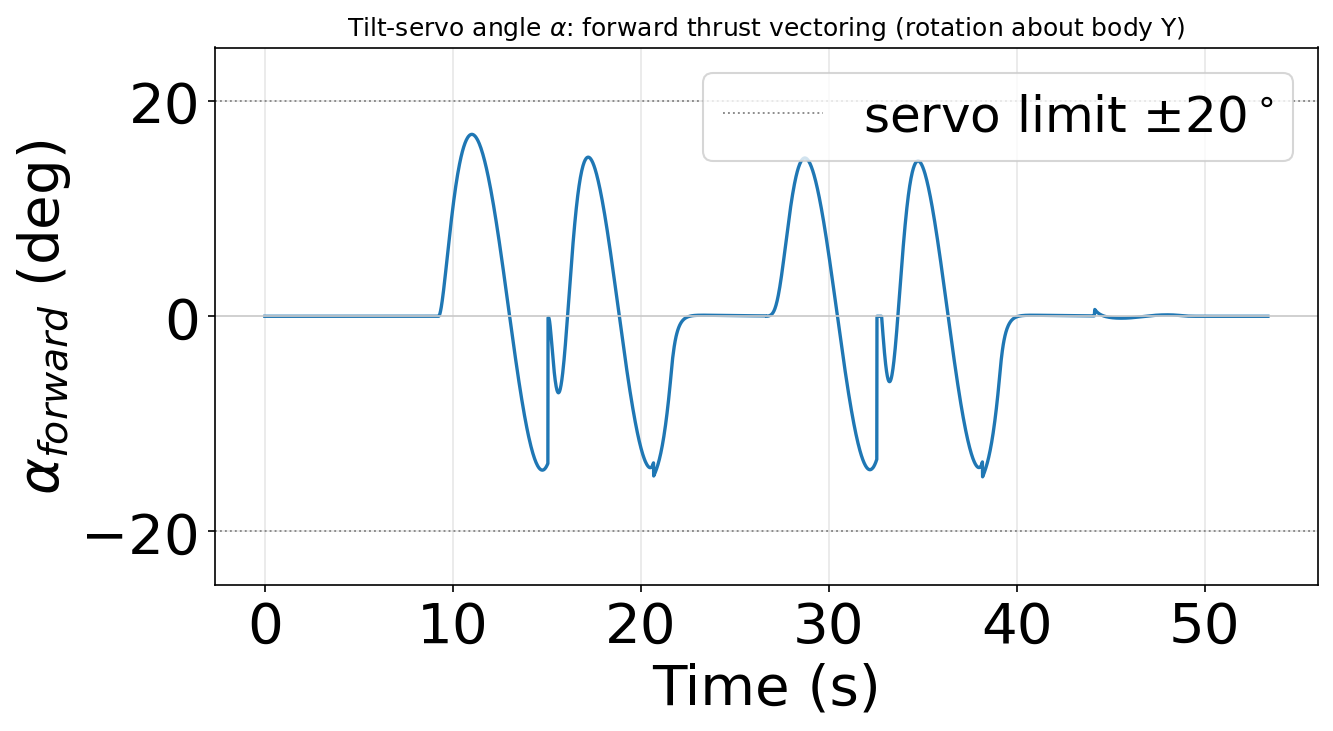}
    \caption{}

\end{subfigure}
\begin{subfigure}[t]{0.16\textwidth}
  \includegraphics[width=\linewidth]{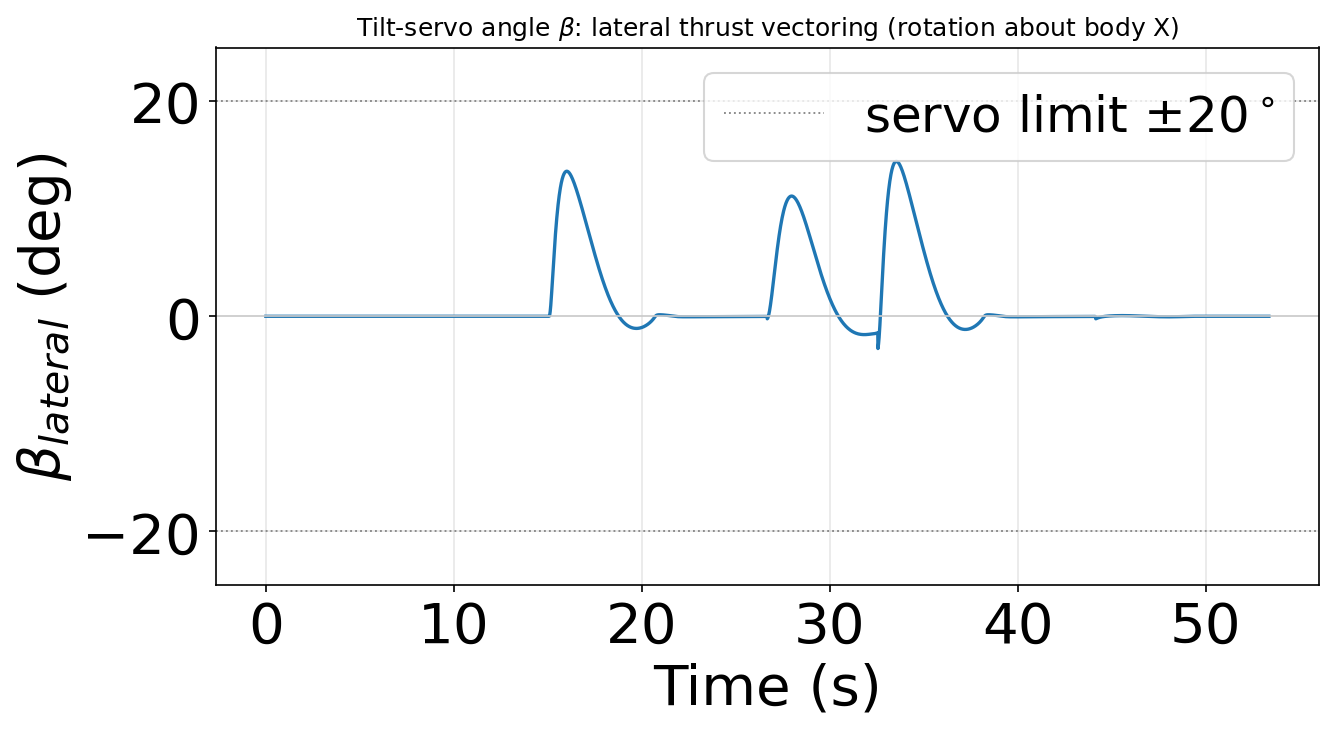}
    \caption{}

\end{subfigure}
\caption{(a) Attitude, (b)-(c) commanded tilt angle alpha \& beta.}
\label{fig:traj}
\end{figure}


\begin{table}[t]
\centering
\caption{Steady-state station-keeping error under wind (m)}
\label{tab:holdstation}
\begin{tabular}{lcc}
\toprule
Controller & 1.5 m/s & 3.9 m/s (OOD)\\
\midrule
PD only & 0.117 & 0.319\\
PD + NN residual & \textbf{0.060} & \textbf{0.162}\\
\bottomrule
\end{tabular}
\end{table}

\section{Conclusion and Limitations}

We present a tilt-rotor quadcopter flight stack with an explicit actuation
model and a cascaded PD controller. State estimation is performed using a
mode-dependent EKF, which switches its fused sensor set through a hysteretic
classifier. A bounded neural residual corrects the steady-state wind offset
of the proportional loop without modifying the controller itself. In
simulation, the EKF reaches a post-settling position RMSE of
\SI{0.226}{\metre} and estimates a \SI{2.3}{\metre\per\second} wind
to within \SI{0.15}{\metre\per\second} per axis, and the residual
reduces steady-state wind error by about \SI{49}{\percent} both within
and beyond its training distribution.

This study has few limitations. The residual is trained by single-round
behavioral cloning from PD-only rollouts, so it has no data from states
visited under its own control (DAgger is the standard remedy). The
out-of-distribution result is a single held-out point, only constant wind
is tested, and the evaluation is simulation-only. The teacher corrects
only the tilt-angle trim, leaving the thrust and torque outputs untested.
Next steps are retraining the network on its own state distribution
(DAgger), testing under gusts and turbulence, and hardware-in-the-loop
and flight validation.

\bibliographystyle{ieeetr}
\bibliography{references}
\end{document}